\documentclass[11pt]{article}
\usepackage[preprint]{acl}
\usepackage{times,latexsym}
\usepackage[T1]{fontenc}
\usepackage[utf8]{inputenc}
\usepackage{microtype,inconsolata}
\usepackage{amsmath,amssymb,graphicx,booktabs,array,tabularx,longtable}
\usepackage{xcolor,float,hyperref,url,placeins}
\definecolor{darkblue}{HTML}{000099}
\newcommand{\sg}{TeachGrade}
\newcommand{\K}{\mathcal{K}}
\newcommand{\R}{\mathcal{R}}
\newcommand{\U}{\mathcal{U}}
\newcommand{\E}{\mathcal{E}}
\newcommand{\pre}{\mathrm{pre}}
\newcommand{\fin}{\mathrm{final}}
\title{TeachGrade: Training-Free Short Answer Scoring with Teacher Evidence}
\author{%
  \parbox{0.97\textwidth}{\centering\normalfont
    \textbf{\large Jinhao Yan\textsuperscript{1}\quad
      Zhihong Shen\textsuperscript{1}\quad
      Shengjie Wang\textsuperscript{1}\quad
      Yiting Ren\textsuperscript{1}}\\
    \textbf{\large Wentao Zhang\textsuperscript{2}\quad
      Yilei Wang\textsuperscript{1}\quad
      Peichao Lai\textsuperscript{2,*}}\\[4pt]
    \fontsize{10}{11.5}\selectfont
    \textsuperscript{1}Fuzhou University\\[2pt]
    \textsuperscript{2}Peking University\\[4pt]
    \fontsize{9}{10.5}\selectfont
    \href{mailto:jinhao.yan@fzu.edu.cn}{\texttt{jinhao.yan@fzu.edu.cn}}\quad
    \textsuperscript{*}\href{mailto:lpc@pku.edu.cn}{\texttt{lpc@pku.edu.cn}}\\[2pt]
    \normalfont\fontsize{9}{10.5}\selectfont
    \textsuperscript{*}Corresponding author.
  }%
}
\hypersetup{pdfauthor={Jinhao Yan, Zhihong Shen, Shengjie Wang, Yiting Ren, Wentao Zhang, Yilei Wang, Peichao Lai}}

\hypersetup{pdftitle={TeachGrade: Training-Free Short Answer Scoring with Teacher Evidence}}
\begin{document}
\maketitle

\begin{abstract}
Multi-subject, multi-step scoring requires consistent partial credit across grading conventions. We introduce \sg{}, a training-free framework using teacher-graded examples. The Task Adapt module builds two frozen memories: Skill captures contextual teacher preferences beyond the rubric; CaseBank supplies local evidence. PreJudge provides retrieval cues; FinalJudge regrades the answer without preliminary scores. On 3,829 SAS-Bench answers across 12 tasks, official macro collaborative consistency score (CCS) rises from Native's 67.99 to 81.32 (+13.33 points), and pooled step exact match from 54.85\% to 62.71\%. Macro quadratic weighted kappa (QWK) reaches 84.71, exceeding Native, OpenJudge, ProofGrader, and Trojsten by 9.52, 3.50, 11.37, and 17.43 points, respectively, and leading on 8 tasks. Local Qwen3.8-27B gains 8.83 CCS and 4.08 QWK points, improving CCS on all 12 tasks. QWK gains over Native span 6.38--14.65 points on Portuguese automatic essay scoring (AES) and 2.16--2.24 on ASAP-SAS across label views, with lower mean absolute error. These results support reuse across languages, response formats, and backbones without weight updates.
\end{abstract}

\section{Introduction}
Short answer scoring (SAS) evaluates a student's response against a question, reference information, and assessment criteria to predict teacher-assigned scores~\citep{mohler2011learning,dzikovska2013semeval}. Reliable automation can reduce repetitive grading work and support educational feedback, especially when assessment requires both an overall score and credit for individual reasoning steps.

\emph{Reliable multi-subject SAS requires fine-grained scoring and adaptation to subject-specific grading conventions.} Fine-grained scoring must preserve valid credit, respect dependencies, and avoid duplicate rewards. Step errors can produce a \emph{cheating-correct result}: over-crediting one step and under-crediting another yields the correct total, although both judgments are wrong. In Figure~\ref{fig:framework}, PreJudge awards 2 instead of 1 for differentiation and 0 instead of 1 for the interval conclusion, yet matches the teacher's total of 4/12. ``Cheating'' denotes misleading aggregate agreement, not intentional manipulation. Across subjects, mathematical reasoning chains and humanities key-point coverage require different grading conventions; transferring one interpretation unchanged can misidentify credit. These fine-grained and adaptation failures undermine diagnostic feedback and consistency across disciplines.

\begin{figure*}[t]
\centering
\includegraphics[width=\linewidth]{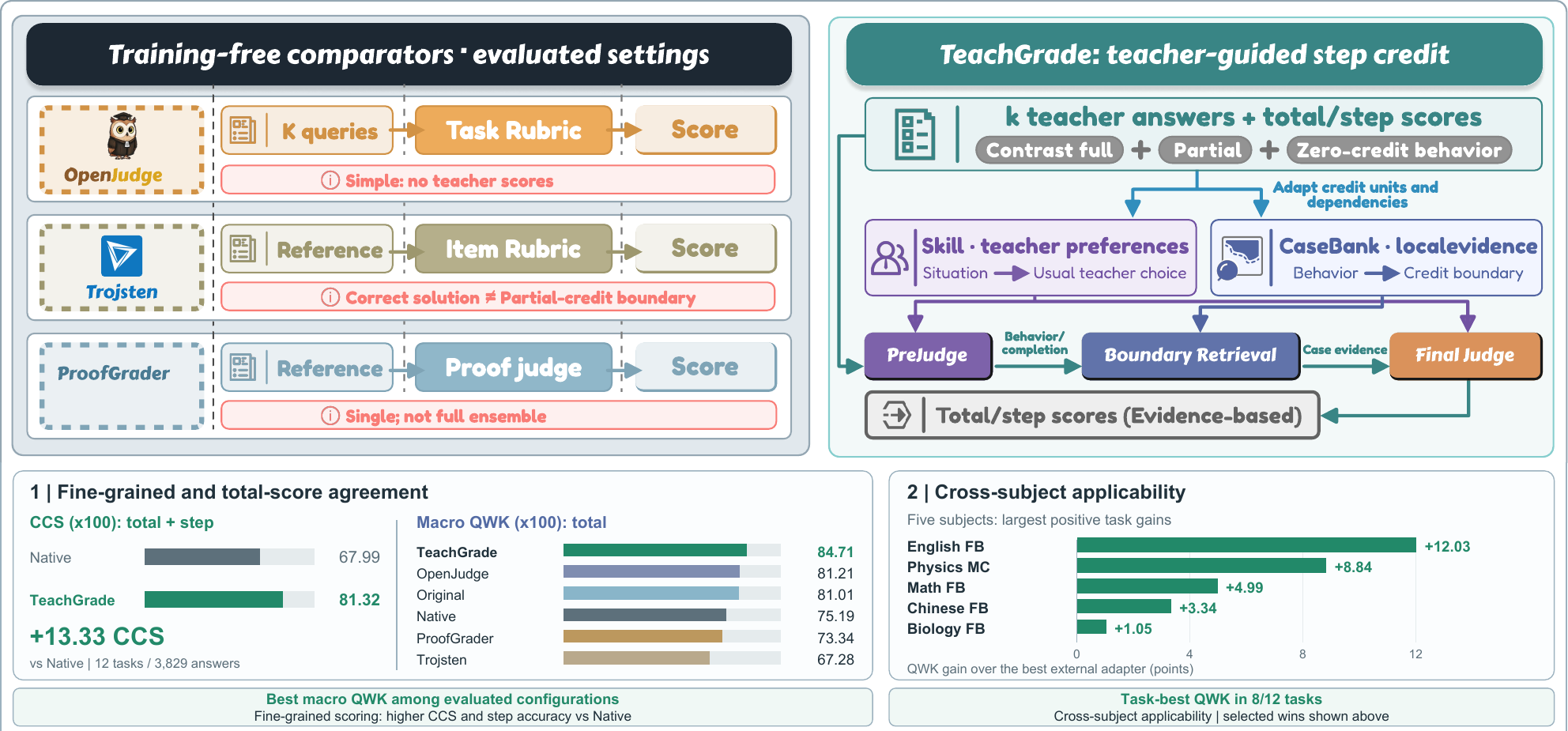}
\caption{\textbf{Fine-grained scoring and cross-subject applicability.} Top: evaluated workflows. Bottom: macro CCS/QWK and five largest subject-wise QWK gains over the best external adapter (one task per subject). Native directly scores current-question inputs without teacher cases, Skill, or retrieval; all use DeepSeek-V4-Flash. Tables~\ref{tab:main}--\ref{tab:macro} cover all 12 tasks.}
\label{fig:comparison}
\end{figure*}

For \emph{fine-grained assessment}, \emph{data-side} supervision extends beyond totals: Portuguese AES scores five writing competencies~\citep{silveira2024portuguese}, while SAS-Bench annotates answer steps and error categories~\citep{lai2026sasbench}. PRM800K supplies human step-correctness labels~\citep{lightman2024verify}; ProcessBench tests first-error identification~\citep{zheng-etal-2025-processbench}. These target complementary aspects: essay competencies, reasoning correctness, and educational partial credit. \emph{Model-side} approaches include rubric-aware content scoring~\citep{wang2019inject}, process-supervised reasoning evaluation~\citep{wang2024mathshepherd}, and ProofGrader's references, marking schemes, and repeated proof judgments~\citep{proofgraderpaper}. Richer annotations and structured judgments enable finer evaluation, but do not determine teachers' partial-credit allocation.

For \emph{cross-subject applicability}, \emph{data-side} resources broaden languages and response forms. SemEval-2013 supports short free-text response analysis with categorical labels~\citep{dzikovska2013semeval}. Portuguese AES adds multi-rater essay evaluation~\citep{silveira2024portuguese}; SAS-Bench spans subject-specific examinations with expert annotations~\citep{lai2026sasbench}. Their scoring units and label granularity differ. \emph{Model-side} adaptation includes LLM-Rubric's learned rater calibration and Rank-Then-Score's fine-tuned Ranker~\citep{hashemi2024llmrubric,cai2025rts}. Alternatively, Trojsten derives item criteria from reference solutions~\citep{zahradnik2025trojsten}; OpenJudge uses descriptions and sampled queries, with an iterative option using labeled preferences~\citep{openjudge}; Auto-Rubric induces criteria from preference examples without weight updates~\citep{xie2025autorubric}. Coverage and adaptable criteria alone do not establish subject-specific partial-credit interpretation.

Gaps remain in \emph{cross-subject adaptation} and \emph{fine-grained grading}: coverage and criteria alone cannot align credit interpretation with disciplinary dependencies and key-point coverage. RTS calibrates essay scores through references, but fine-tunes both Ranker and Scorer with enriched features~\citep{cai2025rts}. OpenJudge's Simple mode generates rubrics without training~\citep{openjudge}, yet descriptions and questions lack contextual partial-credit decisions. Teacher-scored cases, especially step labels, expose these decisions and subject conventions. For retrieval-based SAS, \citet{wang2024generative} train label-aware embeddings to select similar whole responses, without explicitly matching steps by completion. These gaps motivate three challenges: \emph{cross-subject applicability} requires adapting credit interpretation (C1); \emph{fine-grained teacher alignment} requires reusable preferences (C2) and comparable local evidence for applying them (C3).

\textbf{C1: Subject Adaptation.} Coverage alone cannot align scoring with subject conventions, such as mathematical dependencies versus humanities key-point coverage. The challenge is to infer these credit structures from teacher cases without retraining.

\textbf{C2: Teacher Preferences.} Criteria cannot enumerate every expression of correct or incomplete understanding. The challenge is to extract reusable situation--decision instructions from teacher behavior, with concrete cases clarifying their application.

\textbf{C3: Comparable Retrieval.} Whole-response matches can hide step-completion differences. The challenge is to retrieve cases comparable in local content and completion, informing partial credit under current-question requirements.

We propose \sg{}, a teacher-case-guided, training-free framework for fine-grained, multi-subject SAS (Figure~\ref{fig:comparison}). Each task reserves $k$ real teacher cases outside testing: student answers with teacher totals and available step credits. To \emph{teach the scorer how teachers grade}, cases supply references and contextual judgments to emulate. The Task Adapt module builds reusable \emph{Skill} preferences and a local-comparison \emph{CaseBank}, aligning judgments with teacher behavior without weight updates.

\textbf{To address C1}, the Task Adapt module uses an LLM to derive subject-specific credit units, dependencies, and memory-construction plans from teacher cases, adapting the representation within a shared workflow. \textbf{To address C2}, Skill encodes recurring teacher choices as conditional instructions; CaseBank grounds their application to diverse responses. Together, general guidance and local evidence complement assessment criteria. \textbf{To address C3}, PreJudge supplies question, step, and behavior/error rationale text plus independently estimated completion. Retrieval searches the same and neighboring completion bands, ranks by textual relevance, and prioritizes distinct teacher sources within a bounded result set, matching both local content and completion.

Experiments support both objectives: +13.33 official macro CCS over Native with higher step exact match, and macro QWK above Native, OpenJudge, ProofGrader, and Trojsten, leading on most tasks. Teacher-input controls support the workflow beyond demonstrations. Portuguese AES, ASAP-SAS, and local Qwen3.8-27B support cross-language, response-format, and backbone reuse (Tables~\ref{tab:external}--\ref{tab:qwen}).

Our contributions are threefold:
\begin{enumerate}
\setlength{\itemsep}{0pt}
\item We introduce \sg{}, a training-free scoring framework whose Task Adapt module converts teacher evidence into subject-specific memory-construction plans, enabling a shared workflow to accommodate different answering and grading conventions.
\item We develop a complementary teacher-evidence representation: conditional Skill instructions capture reusable judgment patterns, while a completion-aware CaseBank supplies comparable local references for whole-answer grading.
\item We demonstrate improved total--step consistency and leading macro total-score agreement on SAS-Bench, supported by teacher-input controls and validation across additional benchmarks and a local scoring backbone.
\end{enumerate}

\section{Related Work}
Rubric-based SAS identifies score-relevant content~\citep{wang2019inject}. G-Eval derives steps from task descriptions and criteria before form-filling assessment~\citep{liu2023geval}. \citet{zheng2023judging} examine ratings, pairwise judgments, human agreement, and position/verbosity biases. OpenJudge, Trojsten, and ProofGrader construct or use criteria and references~\citep{openjudge,zahradnik2025trojsten,proofgraderpaper}; \sg{} adds teacher-derived instructions and cases.

JudgeLM and Auto-J learn assessment from evaluation data~\citep{zhu2023judgelm,li2023autoj}. Prometheus learns rubric-conditioned evaluation; Prometheus 2 supports custom criteria and rating/comparison formats~\citep{kim2024prometheus,kim2024prometheus2}. LLM-Rubric learns rater calibration; RTS fine-tunes a Ranker to inform a Scorer~\citep{hashemi2024llmrubric,cai2025rts}. \sg{} instead compiles external teacher memories for a frozen scorer.

Retrieval supplies external evidence~\citep{lewis2020rag} and similar demonstrations for in-context prediction~\citep{liu2022examples}. Content and order affect demonstration learning~\citep{min2022demonstrations,lu2022ordered}; long contexts are also position-sensitive~\citep{liu2024lost}. Learned prompt retrieval trains a selector~\citep{rubin2022retrieve}; retrieval-based SAS uses semantically similar responses~\citep{wang2024generative}. \sg{} matches step records by text and completion, capturing content and local progress.

Process supervision evaluates reasoning~\citep{lightman2024verify,wang2024mathshepherd}; educational partial credit also allocates points across contributions and dependencies. Unlike path aggregation or feedback-driven revision~\citep{wang2023selfconsistency,madaan2023selfrefine}, our second judgment receives Skill and teacher evidence, not preliminary scores. Reliability studies motivate checking local accuracy alongside total agreement~\citep{tan2025judgebench,xu2025contextual,jung2025trust,dorner2025limits}.

\section{Method}
\begin{figure*}[t]
\centering
\includegraphics[width=\linewidth]{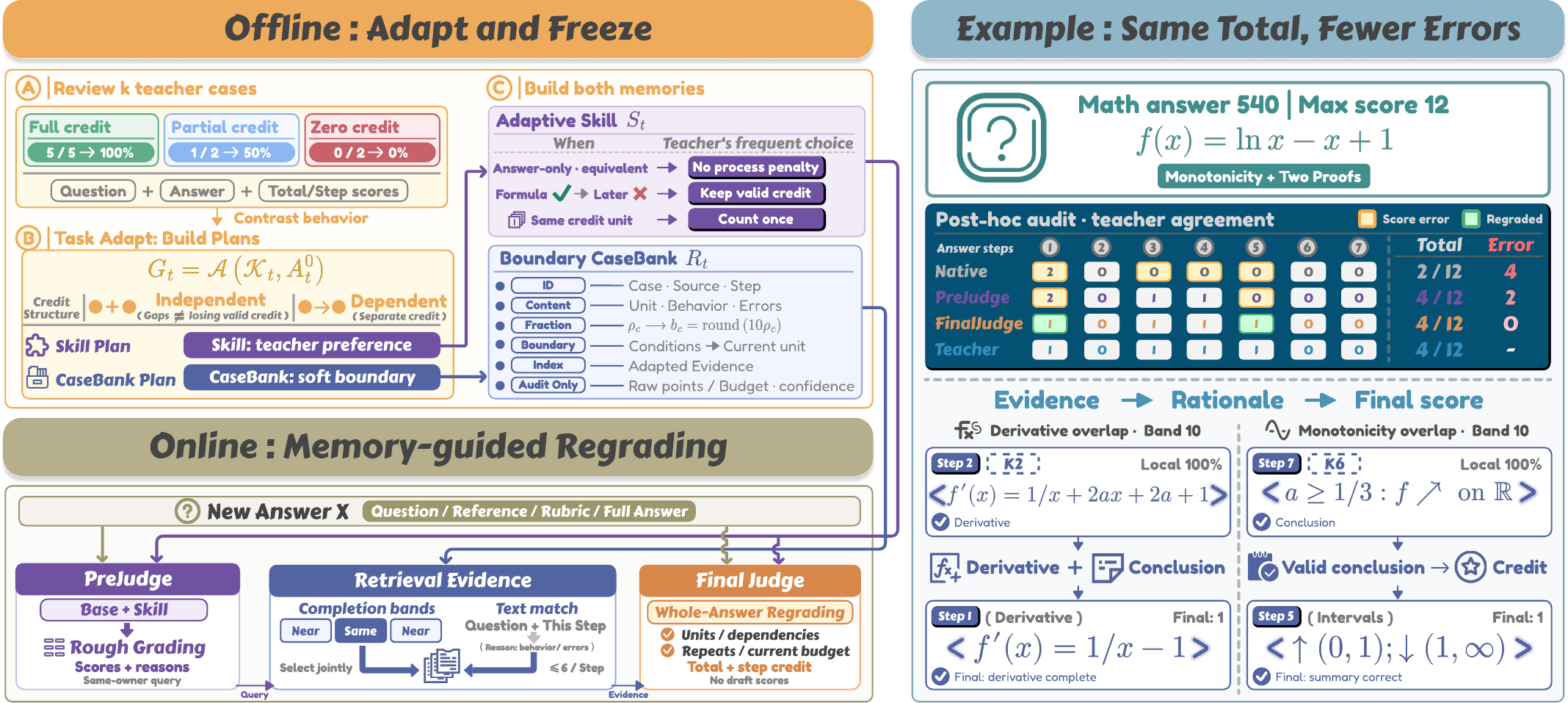}
\caption{\textbf{Offline adaptation and online evidence-guided grading.} The Task Adapt module specifies both memories; PreJudge guides retrieval; FinalJudge regrades without preliminary scores. Native directly scores current-question inputs without teacher cases, Skill, or retrieval. Model outputs use DeepSeek-V4-Flash. The mathematics example improves step credit at an unchanged total; teacher labels are post-hoc audit data.}
\label{fig:framework}
\end{figure*}

\sg{} separates task adaptation from scoring (Figure~\ref{fig:framework}). Offline, the Task Adapt module infers subject-specific plans (C1: Subject Adaptation) and compiles Skill (C2: Teacher Preferences) and CaseBank. Online, PreJudge guides evidence selection (C3: Comparable Retrieval); FinalJudge combines both memories with the current question for whole-answer grading. Memories and model weights remain frozen online.

\subsection{Task and scoring representation}
Fine-grained scoring must distinguish written steps from credited contributions: repeated statements must not earn duplicate credit, nor later errors erase independently valid work. Separating current requirements from teacher observations lets a shared workflow accommodate different grading conventions.

\paragraph{Problem Definition.}
Input $x=(q,r,b,a)$ contains the question, reference information, current rubric and budget, and complete answer. For $m$ annotated steps, the scorer predicts total $\hat y$ and vector $\hat{\boldsymbol{s}}=(\hat s_1,\ldots,\hat s_m)$. Task $t$ has teacher pool
\[
\begin{aligned}
\K_t&=\{(x_i,y_i,\boldsymbol{s}_i)\}_{i=1}^{k},\\
\boldsymbol{s}_i&=(s_{i1},\ldots,s_{i m_i}),
\end{aligned}
\]
where $m_i$ counts teacher answer $i$'s annotated steps. Hats denote predictions; unaccented $y_i$ and $\boldsymbol{s}_i$ are teacher labels. Teacher and test sets $\K_t,\E_t$ have disjoint answer IDs; test scores and error labels are never inputs. Without step labels, teacher evidence remains at answer or assessment-dimension level.

\paragraph{Credit Representation.}
A \emph{credit owner} is a semantic contribution, not a step index: one step may contain several contributions, or several statements repeat one. These units form $\U_x$, with current-question budgets $B_x(u)$. Guided by memory, the Judge interprets units, dependencies, and duplication scope from the current question, rather than using a separate scoring ledger or hard retrieval filter (Appendix~\ref{app:formalrepresentation}).

\subsection{Task adaptation from teacher evidence}
\label{sec:adapt}
To address \textbf{C1: Subject Adaptation}, a shared scorer must adapt its credit interpretation, not just the subject name. Mathematics emphasizes valid operations and dependencies; humanities emphasize key points, concepts, and perspective coverage. Appendix~\ref{app:original} supports this motivation: mathematics-oriented memories transfer unevenly to humanities, while task adaptation improves humanities total-score agreement and Chinese short-answer total--step consistency.

The Task Adapt module uses an LLM to analyze teacher examples $\K_t$ together with original memories $A_t^0$. It produces a Task Specification $G_t$, which is then compiled into Skill $S_t$ and CaseBank $\R_t$:
\begin{equation}
\begin{aligned}
G_t&=\mathcal A(\K_t,A_t^0),\\
(S_t,\R_t)&=\mathcal C(G_t;\K_t,A_t^0),\qquad \theta=\theta_0.
\end{aligned}
\label{eq:adapt}
\end{equation}
Here $\mathcal A$ induces specifications, $\mathcal C$ compiles memories, and $\theta_0$ is the unchanged backbone: adaptation changes external memories, not weights.

\paragraph{Induce Construction Plans.}
The module contrasts higher- and lower-credit behavior within questions or comparable units, identifying rewarded content, accepted equivalences, dependencies, and repeated-credit scope. Its specification contains a Skill plan for situation--decision instructions and a CaseBank plan for retrieval content and completion boundaries. Required-field, source-reference, and coverage checks precede compilation.

\paragraph{Compile Both Memories.}
The Skill compiler organizes and deduplicates preferences, then combines them with original instructions. The CaseBank compiler adds credit-owner information, task-specific retrieval text, and relative completion boundaries, preserving identities and teacher labels. Both outputs are frozen. Shared interfaces encode subject-specific content: principles and applications in politics, versus equivalence, valid operations, and derivation structure in mathematics. Appendix~\ref{app:workedconstruction} provides actual inputs, a specification, and complete memory entries.

\begin{table}[t]
\centering\small

\begin{tabularx}{\linewidth}{@{}l>{\raggedright\arraybackslash}X@{}}
\toprule
Information & Content and online role\\
\midrule
Rubric & Current-question requirements, conditions, and points; governs both judgments.\\
Skill $S_t$ & Teacher-derived condition--action preferences; shared guidance for both Judges.\\
CaseBank $\R_t$ & Local teacher observations of completion and boundaries; evidence for FinalJudge.\\
\bottomrule
\end{tabularx}
\caption{Complementary scoring requirements and teacher memories.}
\label{tab:memory}
\end{table}

\subsection{Skill: conditional teacher-grading preferences}
\label{sec:memory}
To address \textbf{C2: Teacher Preferences}, guidance must cover situations beyond fixed criteria: alternative expressions, incomplete applications, or later errors following correct contributions. The scorer needs reusable instructions capturing teachers' usual choices and their applicability conditions.

Skill represents \emph{situation--decision preferences with an applicability scope}. The Task Adapt module contrasts teacher-scored behavior; the compiler deduplicates preferences, groups them by grading function, and combines them with task semantics and boundaries. Source IDs remain for audit. Unlike rubric requirements and budgets, Skill guides how teachers interpret evidence when applying them (Table~\ref{tab:memory}).

Both Judges receive the same Skill, reusing teacher behavior without prompting with the full pool. CaseBank adds comparable observations for the current response. Appendix~\ref{app:workedconstruction} provides a complete source-linked Skill; Section~\ref{sec:setup} specifies compilation limits.

\subsection{CaseBank-based retrieval-augmented grading}
\label{sec:inference}
To address \textbf{C3: Comparable Retrieval}, references must match both local content and completion: the same topic can involve very different reasoning states. We therefore retrieve within completion bands and rank by question, step, and grading-rationale text.

\paragraph{Normalize Local Evidence.}
Each CaseBank record retains source/step IDs, a credit owner, behavior, errors, completion, and higher/lower boundary evidence. The Task Adapt module enriches its retrieval text with task-specific grading semantics. To compare records across point scales, the compiler estimates the source budget $B_c$ as the highest positive teacher score among cases sharing the question and reference-node text. For local teacher score $s_c$, define
\begin{equation}
\begin{aligned}
\rho_c&=Q_{\delta_s}\!\left(\operatorname{clip}\!\left(\frac{s_c}{B_c},0,1\right)\right),\\
\bar\rho_c&=Q_{\delta_v}(\rho_c),\quad
\widetilde s(c,u;x)=\bar\rho_c B_x(u).
\end{aligned}
\label{eq:ratio}
\end{equation}
Here $Q_\delta$ rounds in $[0,1]$ at resolution $\delta$; $\delta_s,\delta_v$ specify stored and visible precision. If $B_c\leq0$, set $\rho_c=0$ and retain the record. The observed budget is not an annotated maximum. Thus $\widetilde s$ is a \emph{soft reference} under the current budget, not an assigned score; historical points, budgets, and confidence remain audit-only.

\paragraph{Form a Grading-Aware Query.}
PreJudge receives the base scoring input and Skill, producing rough scores, behavioral summaries $a_j^{\pre}$, error descriptions $e_j^{\pre}$, and independently estimated completion $p_j\in[0,1]$. For step text $v_j$, query $q_j$ combines the question, step, and rationale. With at most $h$ references per step,
\begin{align}
z^{\pre}&=J_{\theta}^{\pre}(x,S_t),\label{eq:pre}\\
q_j&=Q(x,v_j,a_j^{\pre},e_j^{\pre}),\notag\\
\R_{x,j}&=\operatorname{Retrieve}_{h}(\R_t,q_j,p_j),\label{eq:retrieve}\\
z^{\fin}&=J_{\theta}^{\fin}\bigl(x,S_t,\{\R_{x,j}\}_{j=1}^{m}\bigr).
\label{eq:final}
\end{align}
The superscripts distinguish prompts, not model parameters. Completion estimates, rather than preliminary points, determine retrieval bands.

\paragraph{Select Comparable Cases.}
Let $L$ partition the unit completion interval into $L$ equal intervals, yielding band indices $0,\ldots,L$. For token sets $T(\cdot)$ and adapted case text $d_c$,
\begin{equation}
\begin{aligned}
b_j&=\operatorname{round}(Lp_j),\quad
b_c=\operatorname{round}(L\rho_c),\\
\operatorname{sim}(q_j,d_c)
&=\frac{|T(q_j)\cap T(d_c)|}
{\sqrt{|T(q_j)|\,|T(d_c)|}}.
\end{aligned}
\label{eq:retrieval-implementation}
\end{equation}
Empty token sets have zero similarity. The first pass takes up to $\ell$ similarity-ranked cases per core band within radius $r$ of $b_j$, prioritizing distinct sources. Below $h$ cases, it fills from remaining core candidates, relaxes source uniqueness, then expands the radius; case IDs stay unique. Completion controls comparability, text controls relevance, and source diversity broadens evidence. Section~\ref{sec:setup} gives precision, bands, and quotas; Appendix~\ref{app:algorithm} specifies tokenization, ties, and safeguards.

\paragraph{Regrade the Whole Answer.}
FinalJudge receives the current input, Skill, and ordered cases, \textbf{not preliminary total or step scores}. It interprets completion under current requirements, considering ownership, dependencies, and duplication across the answer. Budgets are prompt instructions, not hard projections (Appendix~\ref{app:finalconstraints}). Each answer uses two principal generation calls and at most $hm$ evidence associations, plus recovery if needed.

\section{Experiments}
\par
\begin{table*}[t]
\centering

\small
\setlength{\tabcolsep}{4pt}
\begin{tabular}{lrrrrrr}
\toprule
Task & $n$ & \shortstack{Native\\\citeyearpar{lai2026sasbench}} & \shortstack{Trojsten\\\citeyearpar{zahradnik2025trojsten}} & \shortstack{ProofGrader\\\citeyearpar{proofgraderpaper}} & \shortstack{OpenJudge\\\citeyearpar{openjudge}} & \sg{} \\
\midrule
Physics S.A. & 100 & 62.98 & 28.44 & 43.30 & 68.05 & \textbf{69.83} \\
History S.A. & 620 & 89.30 & 86.54 & \textbf{90.67} & 90.18 & 90.07 \\
Physics M.C. & 90 & 82.19 & 38.35 & 80.34 & 83.88 & \textbf{92.72} \\
Geography S.A. & 120 & 62.55 & 64.87 & 59.18 & 61.78 & \textbf{64.94} \\
Biology F.B. & 556 & 89.43 & 83.34 & 87.21 & 93.08 & \textbf{94.13} \\
Chinese F.B. & 119 & 90.93 & 85.98 & 77.77 & 94.14 & \textbf{97.48} \\
Chinese S.A. & 578 & \textbf{79.97} & 70.71 & 55.11 & 79.08 & 78.72 \\
Math S.A. & 781 & 81.17 & 82.76 & 79.59 & 83.12 & \textbf{86.14} \\
Politics S.A. & 200 & 91.80 & 88.18 & 91.07 & \textbf{94.28} & 92.24 \\
English F.B. & 225 & 47.65 & 44.45 & 44.94 & 60.49 & \textbf{72.52} \\
Math F.B. & 410 & 37.07 & 44.39 & 84.14 & 77.82 & \textbf{89.13} \\
Chemistry F.B. & 30 & 87.25 & \textbf{89.39} & 86.81 & 88.62 & 88.59 \\
\midrule
Avg. & 3829 & 75.19 & 67.28 & 73.34 & 81.21 & \textbf{84.71} \\
\bottomrule
\end{tabular}
\caption{Task-level SAS-Bench QWK. Baselines: Native~\citep{lai2026sasbench}, Trojsten~\citep{zahradnik2025trojsten}, ProofGrader~\citep{proofgraderpaper}, and OpenJudge~\citep{openjudge}. Bold indicates the highest point estimate per row. Avg. is the unweighted task mean. Question types are defined in Section~\ref{sec:setup}; baseline configurations in Section~\ref{sec:baselines}.}
\label{tab:main}
\end{table*}
\par

\begin{table}[t]
\centering\small

\begin{tabular}{lrr}
\toprule
Metric & Native & \sg{}\\
\midrule
CCS $\uparrow$ & 67.99 & \textbf{81.32}\\
QWK $\uparrow$ & 75.19 & \textbf{84.71}\\
MAE $\downarrow$ & 1.548 & \textbf{1.248}\\
RMSE $\downarrow$ & 2.301 & \textbf{1.907}\\
Exact $\uparrow$ & 37.61 & \textbf{45.92}\\
\bottomrule
\end{tabular}
\caption{Equal-task averages for Native and \sg{}. MAE, RMSE, and exact match refer to totals; CCS uses the official implementation.}
\label{tab:macro}
\end{table}

We test fine-grained scoring, cross-subject transfer, and reuse across response formats and backbones, presenting settings and baselines before results and analyses.

\subsection{Experimental Setup}
\label{sec:setup}
\paragraph{Datasets.}
We test multi-subject grading on 3,829 SAS-Bench answers in 12 tasks~\citep{lai2026sasbench}: \emph{short answer} (S.A.), \emph{fill-in-the-blank} (F.B.), and \emph{multiple choice} (M.C.). Each task reserves 20 teacher-scored answers for memory construction, excluded from every method's test cohort and metrics, with zero teacher/test answer-ID overlap. Portuguese AES~\citep{silveira2024portuguese} and English ASAP-SAS~\citep{asap2012} test other languages and response forms; Portuguese labels cover competencies and totals, not reasoning steps. Appendix~\ref{app:protocol} and the Supplementary Audit document split manifests, the updated mathematics cohort, and provenance.

\paragraph{Backbones and Hyperparameters.}
Main experiments use DeepSeek-V4-Flash at Judge temperature 0.1 (API provenance: Appendix~\ref{app:protocol}); local validation uses frozen \texttt{Qwen/Qwen3.8-27B} via vLLM on the same split. With $k=20$, Skill retains at most three deduplicated incremental rules per field. Completion has stored resolution $\delta_s=10^{-4}$ and visible resolution $\delta_v=10^{-2}$, serialized as integer percentages but expressed as normalized fractions here. Retrieval uses $L=10$ intervals, radius $r=1$, quota $\ell=2$ per band, and at most $h=6$ cases per step. These evaluated settings are not theoretically optimal thresholds. \emph{Training-free} excludes backbone updates during adaptation and scoring, not teacher supervision.

\paragraph{Evaluation Metrics.}
Official SAS-Bench \emph{collaborative consistency score} (CCS), measuring joint total/step agreement, is primary. \emph{Quadratic weighted kappa} (QWK) measures ordinal total agreement, enabling comparisons with total-only baselines. It adjusts for score marginals and weights larger errors more heavily, rather than measuring accuracy. We also report mean absolute error (MAE), root mean squared error (RMSE), and exact match. QWK, CCS, and exact match are scaled by 100. \emph{Avg.} denotes equal-task means; pooled step metrics weight steps equally. Paired intervals use 3,000 task-stratified question-cluster bootstrap resamples (seed 42); Appendix~\ref{app:metricdetails} gives definitions and audits.

\subsection{Baselines}
\label{sec:baselines}
To compare training-free workflows, baselines share the backbone and current-question inputs. \textbf{Native} uses SAS-Bench's direct prompt without demonstrations~\citep{lai2026sasbench}. \textbf{Trojsten} derives an item rubric from the reference solution~\citep{zahradnik2025trojsten}. \textbf{OpenJudge} uses Simple with the same teacher-pool questions, excluding answers and grades~\citep{openjudge}. \textbf{ProofGrader} uses the official reference-only Single configuration~\citep{proofgraderpaper}: its prompt formatter and reference template, without problem-specific marking schemes or ensembling. Names denote these evaluated configurations; Appendix~\ref{app:protocol} details adapters. Demonstration and supervision controls also receive teacher answers and scores (Section~\ref{sec:analysis}).

\subsection{Main Results}
\par
\begin{table}[t]
\centering

\small
\setlength{\tabcolsep}{2pt}
\begin{tabular}{@{}>{\raggedright\arraybackslash}p{2.2cm}rrrrr@{}}
\toprule
Setting & QWK & CCS & MAE & RMSE & Exact \\
\midrule
Full \sg{} & \textbf{80.18} & \textbf{75.39} & \textbf{1.4448} & \textbf{2.2006} & \textbf{32.66} \\
Raw teacher few-shot & 73.33 & 66.89 & 1.8042 & 2.6259 & 25.54 \\
Retrieved examples & 76.03 & 70.34 & 1.6240 & 2.4178 & 28.01 \\
Without rubric & 78.07 & 74.66 & 1.5864 & 2.3647 & 30.76 \\
Total-only teacher labels & 75.39 & 74.82 & 1.5394 & 2.3333 & 27.91 \\
\bottomrule
\end{tabular}
\caption{Input and supervision controls on four tasks and 1,306 answers. Full is the complete \sg{}\ workflow; task-level results and call counts appear in the appendix.}
\label{tab:ablation}
\end{table}
\par

\label{sec:results}
We test fine-grained and cross-subject grading under the fixed protocol (Tables~\ref{tab:main}--\ref{tab:macro}), prioritizing CCS and using QWK to compare total-only external baselines.

\paragraph{Fine-Grained Agreement.}
Against Native, CCS rises from 67.99 to 81.32 (+13.33), with gains in all five aggregate metrics (Table~\ref{tab:macro}). Step exact match rises from 54.85\% to 62.71\%, while MAE falls from 0.8678 to 0.6012. Question-cluster 95\% intervals are $[6.95,8.79]$ points for exact-match gain and $[0.2419,0.2920]$ for MAE reduction (Appendix~\ref{app:reviewaudit}), showing better step credit alongside total agreement.

\paragraph{Cross-Subject Performance.}
\sg{} has the highest average QWK (84.71), exceeding Native, OpenJudge, ProofGrader, and Trojsten by 9.52, 3.50, 11.37, and 17.43 points, respectively. It leads on 8/12 tasks, gaining up to 12.03 points over the best external configuration. Gaps to the best method are 0.60 on History S.A., 0.80 on Chemistry F.B., 1.25 on Chinese S.A., and 2.04 on Politics S.A.; these modest shortfalls preserve its lead in average QWK and task wins.

\subsection{Analysis and Discussion}
\begin{table}[t]
\centering\small
\setlength{\tabcolsep}{3pt}
\begin{tabularx}{\linewidth}{@{}>{\raggedright\arraybackslash}Xrrrr@{}}
\toprule
Dataset/view & $n$ & QWK & Gain & MAE\\
\midrule
Portuguese all-raters & 216 & \shortstack[r]{57.69\\\textbf{64.07}} & +6.38 & \shortstack[r]{173.704\\\textbf{132.963}}\\
\addlinespace[2pt]
Portuguese median-gold & 72 & \shortstack[r]{58.39\\\textbf{73.04}} & +14.65 & \shortstack[r]{158.889\\\textbf{104.444}}\\
\addlinespace[2pt]
ASAP-SAS Score1 Avg. & 1504 & \shortstack[r]{52.74\\\textbf{54.90}} & +2.16 & \shortstack[r]{0.401\\\textbf{0.380}}\\
\addlinespace[2pt]
ASAP-SAS Score2 Avg. & 1504 & \shortstack[r]{50.94\\\textbf{53.18}} & +2.24 & \shortstack[r]{0.424\\\textbf{0.393}}\\
\bottomrule
\end{tabularx}
\caption{Cross-benchmark comparisons. Each metric cell lists Native above \sg{}; bold marks the better value. Native scores directly without teacher memories. Portuguese all-raters contains 216 ratings of 72 essays. Gain is the QWK difference.}
\label{tab:external}
\end{table}

\par
\begin{table}[t]
\centering

\small
\setlength{\tabcolsep}{2pt}
\begin{tabular}{@{}lrrrrr@{}}
\toprule
Task & $n$ & \multicolumn{2}{c}{QWK} & \multicolumn{2}{c}{CCS} \\
\cmidrule(lr){3-4}\cmidrule(lr){5-6}
 & & Native & \shortstack{Teach\\Grade} & Native & \shortstack{Teach\\Grade} \\
\midrule
Physics S.A. & 100 & 63.60 & \textbf{66.07} & 45.09 & \textbf{60.83} \\
History S.A. & 620 & 91.54 & \textbf{92.03} & 83.29 & \textbf{91.07} \\
Physics M.C. & 90 & 84.95 & \textbf{95.99} & 82.60 & \textbf{97.03} \\
Geography S.A. & 120 & 60.75 & \textbf{61.01} & 55.72 & \textbf{59.41} \\
Biology F.B. & 556 & 91.91 & \textbf{95.00} & 75.82 & \textbf{91.12} \\
Chinese F.B. & 119 & 98.11 & \textbf{99.84} & 93.46 & \textbf{98.45} \\
Chinese S.A. & 578 & 77.18 & \textbf{77.58} & 67.80 & \textbf{74.60} \\
Math S.A. & 781 & 82.69 & \textbf{86.91} & 67.29 & \textbf{82.59} \\
Politics S.A. & 200 & 91.56 & \textbf{92.94} & 75.67 & \textbf{84.08} \\
English F.B. & 225 & 53.02 & \textbf{74.02} & 69.59 & \textbf{77.39} \\
Math F.B. & 410 & 88.03 & \textbf{92.95} & 88.47 & \textbf{93.62} \\
Chemistry F.B. & 30 & \textbf{92.72} & 90.64 & 80.06 & \textbf{80.55} \\
\midrule
Avg. & 3829 & 81.34 & \textbf{85.42} & 73.74 & \textbf{82.56} \\
\bottomrule
\end{tabular}
\caption{Local Qwen3.8-27B on the same evaluation cohort. Native is direct base-model scoring without teacher memories. Bold marks the better value in each metric pair; Avg. gives the equal-task mean.}
\label{tab:qwen}
\end{table}
\par

\label{sec:analysis}
\paragraph{Teacher-Input and Supervision Controls.}
To test structured teacher evidence against demonstrations, we compare inputs on 1,306 fixed answers (Table~\ref{tab:ablation}). Raw Teacher Few-shot uses five score-stratified examples; Retrieved Examples selects five from the same pool by text similarity and contrastive score bands. Both use one Judge, without Skill or PreJudge. Rubric removal retains both Judges and memories; total-only supervision rebuilds memories without teacher step scores or errors.

Full exceeds raw few-shot by 8.50 CCS and 6.85 QWK points, and retrieved examples by 5.05 and 4.15. Rubric removal and total-only labels reduce average QWK by 2.11 and 4.79. These complete-configuration contrasts support combining current criteria, reusable preferences, and local evidence; they compare one-call demonstrations against two-call memory workflows. Appendix~\ref{app:ablation} reports all tasks; Supplementary Audit S.1 retains development controls.

\paragraph{Adaptation and Regrading.}
To examine credit reassignment, we compare mathematics-oriented versus adapted memories and the two online judgments. Adaptation raises humanities total agreement above Native and Chinese short-answer CCS above Original (Appendix~\ref{app:original}). PreJudge--FinalJudge step exact match rises from 60.09\% to 62.71\% (question-cluster gain interval $[1.89,3.37]$ points). In Figure~\ref{fig:framework}, both predict 4/12, yet step errors fall from two to zero; Native predicts 2/12 with four errors. Equal totals thus still require step-level assessment (paired audits: Appendix~\ref{app:audit}).

\paragraph{Teacher-Pool Size.}
To test teacher-budget sensitivity, we compare $k=10,20,30,40$ on one mathematics development cohort, separate from the final evaluation. All outperform Native in CCS and QWK; $k=20$ leads on both. More examples do not monotonically improve scoring, motivating budget checks (Appendix~\ref{app:knowledge}).

\paragraph{Other Languages and Response Formats.}
To test transfer beyond SAS-Bench's formats and step labels, we evaluate Portuguese AES and English ASAP-SAS (Table~\ref{tab:external}). Portuguese QWK gains over Native are 6.38 across raters and 14.65 with median labels, both on the same 72 essays. ASAP-SAS gains 2.16 and 2.24 average QWK points across its label sets. MAE decreases in all four views, supporting answer- or competency-level memories without reasoning-step labels.

\paragraph{Local-Backbone Validation.}
To test local reuse, we evaluate frozen Qwen3.8-27B on the same cohort (Table~\ref{tab:qwen}). CCS rises from 73.74 to 82.56 and QWK from 81.34 to 85.42; unrounded gains are 8.83 and 4.08. CCS improves on all 12 tasks, QWK on 11. Only Chemistry F.B. loses QWK (2.08 points); its 30 answers can be sensitive to individual predictions. All-task CCS gains support reuse across API-served and local scorers.

\section{Conclusion}
\sg{} supports multi-step, multi-subject SAS without weight updates. The Task Adapt module encodes subject-specific conventions in construction plans; Skill captures contextual teacher choices; CaseBank provides content- and completion-comparable references. FinalJudge combines both memories with the current rubric, regrading the whole answer without preliminary scores. This unites task-appropriate grading, reusable preferences, and concrete partial-credit evidence.

The main result is +13.33 CCS over Native with better step accuracy. Macro QWK exceeds OpenJudge, ProofGrader, and Trojsten, leading on 8/12 tasks. Local Qwen3.8-27B gains 8.83 CCS and 4.08 QWK, with all-task CCS gains. Portuguese AES and ASAP-SAS improve QWK and MAE across label views, extending results to essay and answer scoring without step labels. Together with teacher-input controls, these findings support reusable teacher memory across settings and backbones.
\label{main:end}

\section*{Limitations}
\label{sec:limitations}
The frozen teacher memories support answer-disjoint, within-task adaptation; uncommon responses or new grading conventions may require rebuilding them. Retrieval can select unsuitable cases when preliminary completion or behavior estimates are inaccurate. Whole-answer regrading improves average step accuracy, but individual revisions can worsen predictions, and prompt-based budget constraints do not guarantee exact allocation consistency. Appendix~\ref{app:audit} retains paired outcomes and allocation residuals. The comparisons evaluate specified workflows rather than equal-call-budget component effects. Consequential educational decisions require teacher review.

\bibliography{verified_references,processbench_source}
\clearpage
\appendix
\setcounter{table}{0}
\renewcommand{\thetable}{A\arabic{table}}
\renewcommand{\theHtable}{appendix.\arabic{table}}
\setcounter{figure}{0}
\renewcommand{\thefigure}{A\arabic{figure}}
\renewcommand{\theHfigure}{appendix.\arabic{figure}}
\section{Supplementary Experiments}
\subsection{Evaluation protocol and reproducibility}
\label{app:protocol}
The final SAS-Bench cohort contains 3,829 answers across 12 tasks and 20 held-out teacher examples per task. The teacher pool is fixed before scoring, is used only for adaptation and reference construction, and is excluded from every method's evaluation IDs and metrics. Math S.A., Politics S.A., and Chemistry F.B. use whole-dataset shuffling with seed 42 and reserve the last 20 answers for the teacher pool. The other nine tasks retain their existing split-manifest IDs. Frozen manifests specify evaluation IDs and order. The later 781-answer mathematics run replaces the earlier mathematics cohort; the other 11 tasks retain their frozen results. Checks cover prediction counts, unique IDs, zero teacher/evaluation intersections, memory hashes, and exclusion of historical absolute scores from final evidence. Recovery uses the same manifest and memories. Separation is at the answer level: different students' answers to the same question may occur across sets; no test-answer labels enter task adaptation or scoring.

The main tables compare complete workflows under the configurations in Section~\ref{sec:setup}. External rating scales are mapped to current-question discrete scores by a fixed conversion. Demonstration controls use one Judge, whereas rubric removal and total-only supervision retain two. Their contrasts test complete configurations with specified teacher inputs, rather than isolated, budget-matched component effects. Portuguese rating views share the same essays; each benchmark's sample size is reported separately.

ProofGrader uses the official \href{https://github.com/euclidgame/proofgrader/tree/b1a7d0610de7ad0a2353ee2aaa3ed329a69be51c}{ProofGrader repository} (commit \texttt{b1a7d0610de7}). The official \texttt{with\_reference\_solution} template is rendered by \texttt{PromptFormatter}. The adapter supplies the question, reference solution, and concatenated student steps, then parses one judgment's XML score $s\in\{0,\ldots,7\}$. For question maximum $B$, the reported total is $\operatorname{round}(Bs/7)$. Teacher examples and evaluation labels are excluded from the prompt; no SAS-Bench step scores are inferred. This is a reference-only Single adaptation to SAS-Bench, not a reproduction of the original paper's full configuration with problem-specific marking schemes and ensembling on ProofBench.

Memories are constructed once per task and reused read-only. An $m$-step answer uses two principal Judge calls and up to $6m$ evidence associations. The four input controls require 1,306, 1,306, 2,612, and 2,612 planned calls, excluding recovery and construction. Saved final-run calls contain 70,291,145 input and 4,324,862 output tokens. The August 2026 provider usage record identifies DeepSeek-V4-Flash; archived requests retain the legacy name \texttt{deepseek-chat}. DeepSeek documented this name as a compatibility alias for V4-Flash's non-thinking mode.\footnote{\url{https://api-docs.deepseek.com/updates/}, entry dated 2026-04-24.} The model-family identification does not pin an immutable backend for each request or establish the complete provenance of Qwen's offline-memory generation. The accompanying Supplementary Audit (ancillary PDF), Section S.6, retains call counts, provenance scope, and consumed-artifact hashes; these are not a complete billing record.

\subsection{Task-level teacher-input and supervision controls}
\label{app:ablation}
Raw Teacher Few-shot uses five score-stratified teacher examples. Retrieved Contrastive Examples selects five from the same pool using textual similarity and low, middle, and high teacher-score bands. Removing the current-question rubric preserves frozen Skill, CaseBank, and both Judges. Total-only supervision hides teacher step scores and error labels during construction and builds answer-level cases.
\par
\begin{table*}[!t]
\centering

\small
\setlength{\tabcolsep}{4pt}
\begin{tabular}{lrrrrr}
\toprule
Task & Full & Raw few-shot & Retrieved & No rubric & Total only \\
\midrule
Physics S.A. & \textbf{69.83} & 60.96 & 65.86 & 62.73 & 68.77 \\
Math S.A. & 86.14 & 85.18 & 85.38 & \textbf{86.79} & 83.72 \\
Politics S.A. & 92.24 & 84.34 & 89.39 & 90.42 & \textbf{93.86} \\
English F.B. & \textbf{72.52} & 62.83 & 63.49 & 72.36 & 55.19 \\
\bottomrule
\end{tabular}
\caption{Task-level QWK for the four-task controls. Bold marks the best value per task.}
\label{tab:ablationqwk}
\end{table*}
\par

\par
\begin{table*}[!t]
\centering

\small
\setlength{\tabcolsep}{4pt}
\begin{tabular}{lrrrrr}
\toprule
Task & Full & Raw few-shot & Retrieved & No rubric & Total only \\
\midrule
Physics S.A. & 60.14 & 52.55 & 58.29 & 59.52 & \textbf{67.39} \\
Math S.A. & 80.73 & 70.99 & 74.36 & \textbf{81.08} & 73.32 \\
Politics S.A. & \textbf{86.30} & 78.34 & 84.27 & 85.42 & 85.70 \\
English F.B. & \textbf{74.37} & 65.68 & 64.45 & 72.61 & 72.85 \\
\bottomrule
\end{tabular}
\caption{Task-level CCS for the four-task controls under the official frozen-order protocol.}
\label{tab:ablationccs}
\end{table*}
\par

The full workflow has the highest four-task macro CCS and QWK (Table~\ref{tab:ablation}). Task-level results retain the variation behind those averages: rubric removal slightly improves Math S.A. QWK, while total-only supervision improves Politics S.A. QWK and Physics S.A. CCS. The controls support complementary information sources across tasks, not an identical contribution within every task.

\subsection{Teacher-pool sensitivity}
\label{app:knowledge}
The mathematics development cohort compares $k=10,20,30,40$ under the same selection procedure. Every setting exceeds Native on this cohort, and $k=20$ achieves the highest CCS and QWK. This supports the chosen teacher budget while showing that adding examples need not improve grading monotonically. Both pool size and composition change along the curve; these results are not pooled with the final mathematics evaluation.
\par
\begin{table*}[!t]
\centering

\small
\setlength{\tabcolsep}{4pt}
\begin{tabular}{lrrrr}
\toprule
System & Knowledge K & QWK $\uparrow$ & CCS $\uparrow$ & MAE $\downarrow$ \\
\midrule
SAS-Bench Native & 0 & 81.26 & 66.63 & -- \\
Adaptive K10 & 10 & 84.93 & 78.16 & 1.307 \\
Adaptive K20 & 20 & \textbf{86.80} & \textbf{80.88} & \textbf{1.163} \\
Adaptive K30 & 30 & 85.91 & 80.60 & 1.179 \\
Adaptive K40 & 40 & 86.03 & 80.64 & 1.219 \\
\bottomrule
\end{tabular}
\caption{Teacher-pool sensitivity on a shared mathematics development cohort, not pooled with the final mathematics evaluation.}
\label{tab:knowledge}
\end{table*}
\par

\subsection{Original-to-adaptive comparison}
\label{app:original}
\begin{figure}[tbp]
\centering
\includegraphics[width=\linewidth]{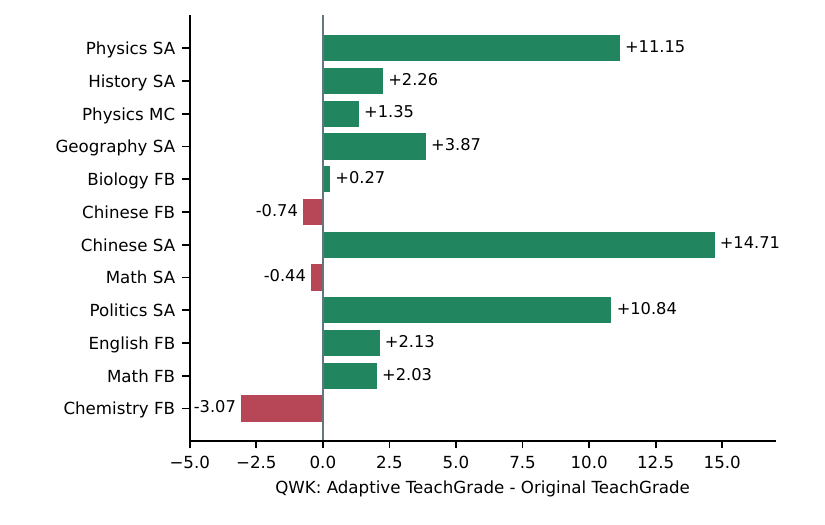}
\caption{Task-level QWK differences between Original and the final adaptive system. Positive and negative values indicate improvement and decline. Values correspond to Table~\ref{tab:originaltasks}.}
\label{fig:originaldelta}
\end{figure}

Original is the static Skill--RAG framework initially developed for mathematical short answers. The final system adds task-specific memory adaptation while retaining the online scoring procedure. Tables~\ref{tab:originaltasks}--\ref{tab:originalgroups} and Figure~\ref{fig:originaldelta} preserve the complete internal comparison, alongside the external comparisons in the main text. The separate representation/consumption development experiment is retained in Supplementary Audit S.1, with all four configurations and task-level outcomes.
\par
\begin{table*}[!t]
\centering

\scriptsize
\setlength{\tabcolsep}{4pt}
\begin{tabular}{lrrrrrrr}
\toprule
Task & $n$ & N QWK & O QWK & A QWK & N CCS & O CCS & A CCS \\
\midrule
Physics S.A. & 100 & 62.98 & 58.68 & \textbf{69.83} & 52.26 & 58.54 & \textbf{60.14} \\
History S.A. & 620 & 89.30 & 87.81 & \textbf{90.07} & 80.98 & 88.20 & \textbf{89.74} \\
Physics M.C. & 90 & 82.19 & 91.37 & \textbf{92.72} & 86.96 & 90.97 & \textbf{93.68} \\
Geography S.A. & 120 & 62.55 & 61.07 & \textbf{64.94} & 56.15 & 60.65 & \textbf{62.85} \\
Biology F.B. & 556 & 89.43 & 93.86 & \textbf{94.13} & 72.74 & 90.02 & \textbf{90.07} \\
Chinese F.B. & 119 & 90.93 & \textbf{98.22} & 97.48 & 86.07 & \textbf{95.78} & 94.85 \\
Chinese S.A. & 578 & \textbf{79.97} & 64.01 & 78.72 & 67.39 & 64.89 & \textbf{74.51} \\
Math S.A. & 781 & 81.17 & \textbf{86.58} & 86.14 & 71.12 & \textbf{82.33} & 80.73 \\
Politics S.A. & 200 & 91.80 & 81.40 & \textbf{92.24} & 70.70 & 80.29 & \textbf{86.30} \\
English F.B. & 225 & 47.65 & 70.39 & \textbf{72.52} & 67.88 & \textbf{74.70} & 74.37 \\
Math F.B. & 410 & 37.07 & 87.10 & \textbf{89.13} & 37.65 & 88.52 & \textbf{89.81} \\
Chemistry F.B. & 30 & 87.25 & \textbf{91.66} & 88.59 & 66.03 & \textbf{85.42} & 78.79 \\
\bottomrule
\end{tabular}
\caption{Complete Native (N), Original (O), and final Adaptive (A) comparisons. Adaptive is the final \sg{}\ system.}
\label{tab:originaltasks}
\end{table*}
\par

\par
\begin{table*}[!t]
\centering

\small
\setlength{\tabcolsep}{4pt}
\begin{tabular}{lrrrrrr}
\toprule
System & QWK $\uparrow$ & CCS $\uparrow$ & MAE $\downarrow$ & RMSE $\downarrow$ & Bias $\to0$ & Exact $\uparrow$ \\
\midrule
Native & 75.19 & 67.99 & 1.548 & 2.301 & \textbf{-0.300} & 37.61 \\
Original \sg{} & 81.01 & 80.03 & 1.460 & 2.194 & -0.818 & 45.70 \\
Adaptive \sg{} & \textbf{84.71} & \textbf{81.32} & \textbf{1.248} & \textbf{1.907} & -0.458 & \textbf{45.92} \\
\bottomrule
\end{tabular}
\caption{Complete Original/final macro comparison. Negative Bias denotes underestimation.}
\label{tab:originalmacro}
\end{table*}
\par

\par
\begin{table*}[!t]
\centering

\small
\setlength{\tabcolsep}{4pt}
\begin{tabular}{lrrrrr}
\toprule
Group & System & QWK $\uparrow$ & CCS $\uparrow$ & MAE $\downarrow$ & RMSE $\downarrow$ \\
\midrule
Open humanities & Native & 80.91 & 68.81 & 1.999 & 2.764 \\
Open humanities & Original & 73.57 & 73.51 & 2.442 & 3.421 \\
Open humanities & Adaptive & 81.49 & 78.35 & 1.945 & 2.687 \\
Structured science & Native & 73.98 & 65.84 & 1.125 & 1.852 \\
Structured science & Original & 91.00 & 88.73 & 0.778 & 1.327 \\
Structured science & Adaptive & 91.14 & 88.09 & 0.744 & 1.305 \\
\bottomrule
\end{tabular}
\caption{Internal development comparison of open-ended humanities and structured-science tasks.}
\label{tab:originalgroups}
\end{table*}
\par

Macro QWK improves by 3.70 points, with a task-stratified answer-level paired bootstrap interval of $[1.67,5.78]$. Open-ended humanities QWK rises from 73.57 to 81.49, while the structured-science mean is nearly unchanged. Math S.A. remains close to Original (86.58 to 86.14); Chemistry F.B. decreases from 91.66 to 88.59. Signed Bias moves from Original's $-0.818$ toward zero ($-0.458$), although Native remains closer at $-0.300$.

In Political\_ShortAns\_79, the student expresses three distinct credit units in different wording. Original assigns 4/4/2; the final system assigns 6/6/3, matching the teacher. Table~\ref{tab:originalrepresentation} relates this example to the adapted representation: the current question supplies the independent budgets, while memory guides interpretation of the credited content.
\par
\begin{table*}[!t]
\centering

\small
\setlength{\tabcolsep}{4pt}
\begin{tabular}{p{.65\linewidth}rrr}
\toprule
Evidence in the student answer & Teacher & Original & Adaptive \\
\midrule
Counters attempts to discredit heroes using sound values, historical facts, and core values & 6 & 4 & \textbf{6} \\
Defends heroes by reconstructing history, identifying misinformation, and promoting the national spirit & 6 & 4 & \textbf{6} \\
Calls for learning from and practicing the spirit of heroes & 3 & 2 & \textbf{3} \\
\midrule
Total & 15 & 10 & \textbf{15} \\
\bottomrule
\end{tabular}
\caption{Step evidence and scores for Political\_ShortAns\_79. Text summarizes the original answer's meaning. Bold highlights corrected scores matching the teacher.}
\label{tab:trace}
\end{table*}
\par

\par
\begin{table*}[!t]
\centering

\small
\setlength{\tabcolsep}{4pt}
\begin{tabular}{>{\raggedright\arraybackslash}p{.17\linewidth}>{\raggedright\arraybackslash}p{.33\linewidth}>{\raggedright\arraybackslash}p{.39\linewidth}}
\toprule
Dimension & Original \sg{} & Adaptive \sg{} \\
\midrule
Credit unit & Generic task-level point-based guidance & Defines \textbf{key principles} in the current reference as \textbf{credit owners} \\
Equivalence & Allows similar meanings without evidence-specific scope & Credits equivalent statements; \textbf{excludes incorrect principles} or confused concepts \\
Contrasts & Cases primarily record the teacher's awarded points & Records \textbf{higher/lower boundaries}, supporting and contrasting cases, and confidence markers \\
Duplication and dependency & Generic Skill warns against duplicate credit & \textbf{Counts repeated owners once}; propagates earlier errors \textbf{only with supporting evidence} \\
Score scale & Historical points may implicitly anchor the current score & Transfers \textbf{completion fractions} under the \textbf{current budget} \\
\bottomrule
\end{tabular}
\caption{Original and adaptive scoring representations; neither changes the question-specific rubric. Bold highlights the adapted representation's key distinctions.}
\label{tab:originalrepresentation}
\end{table*}
\par

\subsection{Step accuracy and paired regrading}
\label{app:audit}
\label{app:reviewaudit}
The audit aligns 3,829 frozen predictions and 13,530 steps by answer ID and checks teacher totals, step labels, and step counts. Total-score metrics average tasks equally; step metrics pool all steps. Table~\ref{tab:auditfull} reports the full PreJudge--FinalJudge comparison. Its intervals resample answers within tasks; Table~\ref{tab:clustersteps} separately reports question-cluster intervals, preserving dependence among responses to the same question.
\begin{table*}[!t]
\centering\small

\begin{tabular}{lrrl}
\toprule
Metric & PreJudge & FinalJudge & Gain and 95\% interval\\
\midrule
Avg. QWK &83.69&84.71&$1.02\ [-0.28,2.35]$\\
Avg. total MAE &1.3378&1.2479&$0.09\ [0.04,0.13]$\\
Avg. total exact match &43.38&45.92&$2.54\ [0.83,4.23]$\\
Pooled step MAE &0.6176&0.6012&$0.0164$ (see Table~\ref{tab:clustersteps})\\
Pooled step exact match &60.09&62.71&$2.62\ [1.87,3.34]$\\
\bottomrule
\end{tabular}
\caption{Full paired regrading audit. Gains are oriented toward improvement; QWK's interval includes zero.}
\label{tab:auditfull}
\end{table*}

\begin{table*}[!t]\centering\small

\begin{tabular}{lrr}\toprule
Contrast & Estimate & 95\% interval\\\midrule
Step exact gain vs. Native & 7.86 & [6.95,8.79]\\
Step MAE reduction vs. Native & 0.2666 & [0.2419,0.2920]\\
Step exact gain vs. PreJudge & 2.62 & [1.89,3.37]\\
Step MAE reduction vs. PreJudge & 0.0164 & [0.0040,0.0293]\\
\bottomrule\end{tabular}
\caption{Question-cluster paired step gains. Positive values indicate improvement by \sg{}.}
\label{tab:clustersteps}
\end{table*}

\begin{table*}[!t]\centering\small

\begin{tabular}{lrrr}\toprule
Task & Native & PreJudge & \sg{}\\\midrule
Physics S.A. & 58.00 / 0.602 & 62.36 / 0.491 & 64.91 / 0.491\\
History S.A. & 30.28 / 1.858 & 31.12 / 1.268 & 34.77 / 1.275\\
Physics M.C. & 67.03 / 0.560 & 86.81 / 0.209 & 82.42 / 0.275\\
Geography S.A. & 44.59 / 0.947 & 44.85 / 0.858 & 48.81 / 0.818\\
Biology F.B. & 60.71 / 1.039 & 72.89 / 0.442 & 69.58 / 0.466\\
Chinese F.B. & 89.86 / 0.125 & 94.49 / 0.067 & 94.78 / 0.061\\
Chinese S.A. & 36.13 / 1.561 & 44.39 / 1.170 & 45.84 / 1.182\\
Math S.A. & 49.57 / 0.695 & 54.11 / 0.581 & 59.80 / 0.533\\
Politics S.A. & 33.08 / 1.162 & 39.41 / 0.902 & 43.23 / 0.845\\
English F.B. & 85.50 / 0.155 & 85.45 / 0.156 & 87.67 / 0.134\\
Math F.B. & 82.59 / 0.817 & 96.43 / 0.129 & 95.76 / 0.172\\
Chemistry F.B. & 50.00 / 0.816 & 52.63 / 0.763 & 51.97 / 0.829\\
\bottomrule\end{tabular}
\caption{All-task step accuracy. Each cell reports exact match / MAE.}
\label{tab:tasksteps}
\end{table*}

Total absolute error improves / stays unchanged / worsens for 728 / 2,581 / 520 answers; corresponding step counts are 1,677 / 10,427 / 1,426. FinalJudge recovers 344 exact totals and changes 218 previously exact totals to incorrect ones, a net gain of 126. Stored totals are unchanged by post-processing; the audit includes one rule-corrected step and eleven rule-corrected preliminary retrieval anchors. Positive and negative revisions remain in every aggregate. The four-task subset is documented separately in Supplementary Audit S.3.

For $r_i=\hat y_i-\sum_j\hat s_{ij}$, the numbers of answers with nonzero residuals are 1,215 / 556 / 277 for Native / PreJudge / \sg{}, compared with 15 in teacher labels. This checks allocation consistency without replacing model totals by step sums. Sign-specific counts are retained in Supplementary Audit S.6.

\section{Construction and Retrieval Details}
\subsection{Task representation and executable workflow}
\label{app:algorithm}
\label{app:formalrepresentation}
\label{app:adapterdetails}
For $x=(q,r,b,a)$, the scorer predicts a total and the annotated step scores. Credit owners $\U_x$ and budgets $B_x(u)$ are interpreted from the current rubric, not identified with step indices: a unit may span statements, and a step may contain several units. Teacher and evaluation answer IDs are disjoint. A total alone can correspond to multiple local allocations, motivating step-labeled teacher evidence where available.

The evaluated V30 adapter receives all 20 teacher records, the complete original Skill, the CaseBank schema and its first 20 representative cases, and the existing Judge interface constraints. Question, reference, rationale, and step text are capped at 1,200, 1,400, 1,200, and 520 characters, respectively; step scores and error labels are retained. Compilation subsequently reads every original case. The workflow is:
\begin{enumerate}
\item Contrast higher- and lower-credit teacher behavior within questions or comparable credit owners. Infer rewarded content, equivalence, dependencies, and duplication scope.
\item Produce a Task Specification with separate \texttt{skill\_compilation} and \texttt{rag\_compilation} plans. Validate required fields, teacher-source IDs, evidence coverage from at least four distinct teacher examples, and model-proposed validation status.
\item Use \texttt{compile\_skill} to deduplicate conditional preferences and prepend them to the original Skill. Use \texttt{compile\_rag} to enrich all original cases with adapted retrieval text and relative completion boundaries. Check that identities and teacher labels are unchanged, then freeze both memories.
\item Run PreJudge with the current answer and Skill. Its separately estimated completion percentages, divided by 100 to give $p_j\in[0,1]$, locate retrieval bands; question, step, and rationale text rank cases within those bands.
\item Select at most six source-linked cases per step and serialize their model-visible evidence. Exclude preliminary scores and historical absolute points.
\item Regrade the entire answer with FinalJudge under the current rubric, shared Skill, and selected evidence. Keep the final model's total without replacing it by a computed step sum.
\end{enumerate}
The formal builder uses structural and provenance checks rather than the numerical support-count threshold of the separate development constructor. That earlier heuristic is documented only in Supplementary Audit S.5.

The frozen selector, \texttt{\_select\_step\_cases\_zh}, lowercases text and constructs token sets from contiguous Chinese sequences of length 1--4, contiguous English letters, and integers/decimals. Equation~\eqref{eq:retrieval-implementation} uses neither term frequencies, embeddings, nor a learned retriever. The evaluated $L=10$ gives bands bounded to 0--10 using Python's ties-to-even \texttt{round}; first-pass diagnostics may list an empty neighboring band outside that range.

Each core band initially contributes at most two similarity-ranked cases, skipping repeated source IDs. Remaining core-band candidates are pooled and ranked, first with distinct sources, then allowing repeated sources if fewer than six cases are selected. Fallback expands by radii 2, 3, and so on with the same source priority. If a positive-band query selects only zero-band cases, an eligible nonzero case replaces the last selection when available in the considered bands or the current/preceding upper band. Case IDs remain unique; exhausted candidates may yield fewer than six cases.

Case compilation divides source-local teacher points by the observed local budget, clips to $[0,1]$, and stores four decimal places. A nonpositive budget yields zero without removing the source. The final evidence view contains case ID, historical completion percentage and band, credit point, behavior, relative boundary, current-budget interpretation, soft-reference guidance, and error type. Historical absolute points, source budgets, and confidence markers remain audit-only; confidence is not a similarity multiplier or score weight.

\phantomsection\label{app:finalconstraints}
FinalJudge is prompted to satisfy
\begin{equation}
\hat s_j\geq0,\qquad 0\leq\sum_{j=1}^{m}\hat s_j\leq\hat y\leq M_x,
\label{eq:budget}
\end{equation}
where $M_x$ is the question maximum. These are prompt-level targets, not hard projections. Actual residuals are reported in Appendix~\ref{app:audit}.

\subsection{Worked specification, Skill, and CaseBank record}
\label{app:workedconstruction}
The Politics S.A. artifacts instantiate Equation~\eqref{eq:adapt}: 20 teacher records produce a task-specific specification, a seven-field Skill, and 95 CaseBank records. The ancillary materials retain complete original-language inputs, specifications, memories, builder, and prompt reconstructions in \texttt{construction\_example/}; \texttt{task\_comparison/} supplies the consumed Math S.A. artifacts. The following English excerpts are explanatory translations, not replacements for serialized records.

For \texttt{Political\_ShortAns\_55}, the question concerns cultural innovation, practice and knowledge, and the value of life in a source passage. The answer maximum is 26 and teacher total 22. Step 5 (stored index 4) says that practical problems prompted the subject to organize and study veterinary knowledge. It receives 2 points and an incomplete-work/omission label. This supplies an observed distinction between stating a valid principle and sufficiently applying it to the material.

\begin{table*}[!t]\centering\small

\begin{tabularx}{\linewidth}{l>{\raggedright\arraybackslash}X}\toprule
Plan component & Stored purpose\\\midrule
Credit semantics & Principles and their application to source material; current-reference credit owners.\\
Evidence and scope & Teacher behavior, higher/lower contrasts, equivalence, dependencies, duplication, and source IDs.\\
Skill plan & Workflow, principles, partial credit, deductions, zero credit, high credit, and boundaries.\\
CaseBank plan & Retrieval goals, owner mapping, behavior/error patterns, dependencies, and relative boundaries.\\
Consumption contract & Completion-based references interpreted under the current budget; historical absolute points hidden.\\
\bottomrule\end{tabularx}
\caption{Construction plan extracted from the actual Politics specification.}
\label{tab:actualspec}
\end{table*}

One complete incremental Skill entry is:
\begin{quote}\small\raggedright
\textbf{Rule:} When the principle is correct but its application to the material is insufficient, award approximately half the credit for that point.\par
\textbf{Evidence IDs:} \nolinkurl{sample:Political\_ShortAns\_55:S4}, \nolinkurl{sample:Political\_ShortAns\_139:S1}, \nolinkurl{sample:Political\_ShortAns\_173:S4}, \nolinkurl{sample:Political\_ShortAns\_35:S4}.\par
\textbf{Confidence marker:} 0.85.
\end{quote}
The compiler selects at most three deduplicated incremental rules per field, combines them with task semantics and boundaries, and prepends them to the original Skill. Evidence IDs and confidence remain in the specification. The instruction expresses a conditional tendency; the current rubric and complete answer determine actual points.

Case \texttt{K1\_S4} comes from the same answer's step index 4. Cases are grouped by question and reference-node text; the maximum positive observed teacher score supplies the local budget. Observations of 2 and 3 give budget 3, fraction $\operatorname{round}_4(2/3)=0.6667$, displayed completion 67\%, and band 7. The budget is observation-derived rather than an independently annotated step maximum.
\begin{table*}[!t]\centering\small

\begin{tabularx}{\linewidth}{l>{\raggedright\arraybackslash}X}\toprule
Layer & Content\\\midrule
Identity & Case/source/step IDs, question summary, complete local answer, reference node.\\
Teacher observation & 2 points; observed budget 3; question maximum 26; incomplete-work label.\\
Retrieval & Fraction 0.6667; band 7; question, step, behavior, reference, error, and adapted goals.\\
Boundary status & Same-question, same-owner contrast; construction marker 1.0.\\
FinalJudge view & 67\%, credit point, behavior, relative boundary, scope, and current-budget interpretation.\\
Audit-only fields & Historical absolute points, source budget, confidence, and original construction records.\\
\bottomrule\end{tabularx}
\caption{Stored and visible layers of the actual case \texttt{K1\_S4}.}
\label{tab:actualcase}
\end{table*}
The 1.0 boundary marker describes the observed contrast group, not a calibrated correctness probability. Of the 95 Politics cases, 24 have actual within-question, same-owner score contrasts; the others retain one-sided status. Replaying both compilation functions reproduces the consumed Skill and all cases field by field. Prompt exports are source-reconstructed and match logged lengths, but historical request-byte equality remains unverified. Full hashes and reconstruction checks are supplied with the companion artifacts.

\subsection{Retrieval coverage and framework example}
\label{app:coverage}
\label{app:fullfigures}
\par
\begin{table*}[!t]
\centering

\scriptsize
\setlength{\tabcolsep}{4pt}
\begin{tabular}{lrrrrrr}
\toprule
Task & K & Cases & Eval. steps & Cases/step & Unique used & Utilization \\
\midrule
Physics S.A. & 20 & 32 & 550 & 6.00 & 31 & 96.88\% \\
Math S.A. & 20 & 101 & 4234 & 6.00 & 97 & 96.04\% \\
Politics S.A. & 20 & 95 & 916 & 6.00 & 88 & 92.63\% \\
English F.B. & 20 & 199 & 2158 & 6.00 & 113 & 56.78\% \\
\bottomrule
\end{tabular}
\caption{Observed retrieval coverage. Case utilization measures breadth of use, not retrieval accuracy.}
\label{tab:coverage}
\end{table*}
\par

Case utilization measures the breadth of consumed records, not retrieval accuracy. The four tasks use multiple sources, with lower utilization in English. The actual query/evidence trace below illustrates how completion bands and textual cues operate together.

In Math\_ShortAns\_540 (Figure~\ref{fig:framework}), Native predicts 2/12; PreJudge and FinalJudge both predict 4/12, with incorrect-step counts 4, 2, and 0 respectively. All seven final step scores match the teacher. Steps 1 and 5 receive preliminary points 2 and 0 but both have 100\% estimated completion, so each uses a band-10 query and retrieves six cases. Replaying all seven queries reproduces selected IDs, order, bands, quotas, and similarities to four decimals.

K2 Step 2 and K6 Step 7 are each second in their respective selection lists. The former exemplifies a complete local derivative; the latter retains a valid monotonicity conclusion despite another unfinished branch. These cases correspond to FinalJudge's separate credit for the derivative and valid interval conclusion. FinalJudge assigns one point to each under the current rubric, without receiving the preliminary points. Dashed links in the figure indicate an explanatory correspondence, not logged causal attribution to a particular retrieved case.

\section{Metrics and Statistical Analysis}
\label{app:metricdetails}
Official SAS-Bench CCS measures joint total/step agreement; QWK measures ordinal total-score agreement. External adapters supply total-score comparisons. QWK, CCS, and exact match use the $\times100$ scale. Macro metrics weight tasks equally; pooled step metrics weight steps equally.

Following the official implementation, scores are converted to integers using the scoring scale, and the observed union of teacher and predicted values is encoded as $L_t$ ordered levels. With joint proportions $O_{ab}$ and $E_{ab}=O_{a+}O_{+b}$,
\begin{equation}
\begin{aligned}
\operatorname{QWK}_t&=1-\frac{\sum_{a,b}w_{ab}O_{ab}}{\sum_{a,b}w_{ab}E_{ab}},\\
w_{ab}&=\frac{(a-b)^2}{(L_t-1)^2}.
\end{aligned}
\label{eq:qwk}
\end{equation}
For $L_t\geq2$, indices range from 0 to $L_t-1$; unobserved intermediate values are not inserted. A zero expected disagreement leaves QWK undefined. Tasks are encoded separately. Aggregation is
\begin{equation}
\begin{aligned}
\operatorname{Avg.}(M)&=\frac1T\sum_{t=1}^{T}M_t,\\
\operatorname{StepMAE}&=\frac{\sum_{i,j}|\hat s_{ij}-s_{ij}|}{\sum_i m_i}.
\end{aligned}
\label{eq:aggregation}
\end{equation}
Step exact match uses the same step denominator.

\phantomsection\label{app:ccsprotocol}
CCS uses the official SAS-Bench implementation at commit \texttt{4ef718b6}, with the function and calling script verified byte for byte.\footnote{\href{https://github.com/PKU-DAIR/SAS-Bench/blob/4ef718b6433623d492d483a38d4045a4b32d6f45/utils/collaborative_consistency_score.py}{Official CCS implementation at the frozen commit}.} Results use the frozen manifest order. The implementation's weight indexing can make order consequential when step counts vary; Supplementary Audit S.4 retains the order-variation tests and alternative-coordinate diagnostics. Those diagnostics do not establish uniform PreJudge--FinalJudge gains across definitions; official CCS remains the paper's benchmark metric.

Question-cluster bootstrap groups whitespace-normalized full question text within tasks, yielding 1,019 clusters. It samples clusters with replacement within each task, retaining all their answers, for 3,000 replicates with seed 42. Every replicate computes task-macro total metrics and pooled step errors; all are defined. Earlier answer-level intervals are explicitly identified in Appendix~\ref{app:audit}. The intervals condition on the evaluated tasks and fixed teacher pools. Supplementary Audit S.4 retains fixed-grid QWK comparisons and their cluster intervals, separately from the official QWK results.

\end{document}